\documentclass[runningheads]{llncs}
\usepackage{graphicx}
\usepackage{amsmath}
\usepackage{algorithm}
\usepackage{algorithmic}
\usepackage{subfigure}
\usepackage{amssymb}
\usepackage{booktabs}
\usepackage{color}
\begin{document}   
\title{Objects as Extreme Points}
\author{Yang Yang\inst{1}\and
Min Li\inst{1} \and Bo Meng\inst{1,2} \and Zihao Huang\inst{1}\and\\
 Junxing Ren\inst{1} \and Degang Sun\inst{1}
}
%
\institute{Institute of Information Engineering, Chinese Academy of Sciences
\and The First Research Institute of The ministry of Public Security
\\
\email{\{yangyang1995, limin, renjunxing, sundegang\}@iie.ac.cn,\\
bit.meng@live.com}}
\maketitle              
\begin{abstract}
    Object detection can be regarded as a pixel clustering task, and its boundary is determined by four extreme points (leftmost, top, rightmost, and bottom). However, most studies focus on the center or corner points of the object, which are actually conditional results of the extreme points. In this paper, we present an Extreme-Point-Prediction-Based object detector (EPP-Net), which directly regresses the relative displacement vector between each pixel and the four extreme points. We also propose a new metric to measure the similarity between two groups of extreme points, namely, Extreme Intersection over Union ($EIoU$), and incorporate this $EIoU$ as a new regression loss. Moreover, we propose a novel branch to predict the $EIoU$ between the ground-truth and the prediction results, and take it as the localization confidence to filter out poor detection results. On the MS-COCO dataset, our method achieves an average precision (AP) of 44.0\% with ResNet-50 and an AP of 50.3\% with ResNeXt-101-DCN. The proposed EPP-Net provides a new method to detect objects and outperforms state-of-the-art anchor-free detectors.

\keywords{Object Detection\and Extreme Points \and Localization \and \\Regression Loss}
\end{abstract}
\section{Introduction}
Object detection is a crucial prerequisite for many computer vision tasks, such as instance segmentation \cite{he2017mask} and multi object tracking \cite{wojke2017simple}. It also plays an essential role in many downstream technologies, such as intelligent video analysis and autonomous driving. Benefiting from the excellent performance of anchors, the detection accuracy of one-stage \cite{redmon2017yolo9000} and two-stage \cite{ren2016faster} object detectors has substantially improved. However, these detectors rely excessively on predefined anchors, thus requiring fine-tuning when training, and lead to poor generalization performance. Anchor-free detectors \cite{Tian_2019_ICCV,kong2020foveabox} have recently drawn much attention for their simple design, great accuracy, and high speed. Generally, anchor-free detectors can be classified into \textbf{key-point-based prediction} and \textbf{dense prediction}.   
\begin{figure}[htbp]
    \centerline{\includegraphics[width=0.62\textwidth]{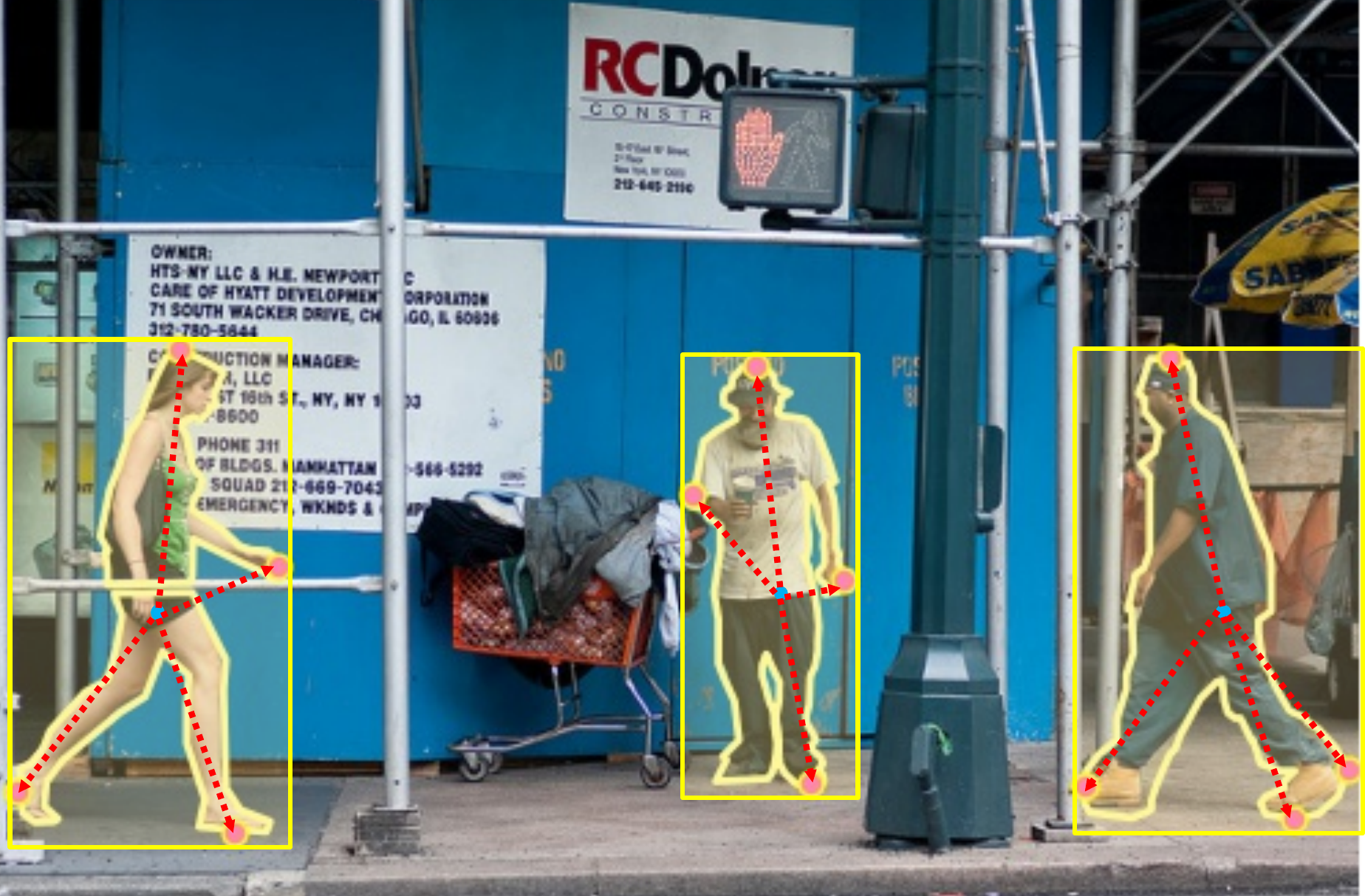}}
    \caption{\textbf{{Illustration of EPP-Net predictions}}. As shown in this image, since the boundary of an object is determined by the extreme points, the bounding box (bbox) is actually a conditional result. Therefore, EPP-Net predicts four relative displacements, an 8D vector as the location of the object.
    }
    \label{example1} 
    \end{figure} 

\subsection{Key-point-based prediction}
The location of an object is usually represented by the smallest enclosing rectangle called the bounding box (bbox). Nevertheless, not all objects can perfectly fit into a rectangle, such as objects with a tilt angle. Therefore, bottom-up methods have been proposed to detect objects in a key-point-based fashion. CornerNet \cite{law2018cornernet} represents an object using a pair of corner points (top left and bottom right), whereas \cite{duan2019centernet} also predicts the center point besides corner points. ExtremeNet \cite{zhou2019bottom} argues that corner points usually lie outside the object and lack appearance features. Therefore, it utilizes the four extreme points and the center points to represent the object. Free from the limitations of the rectangular box, these key-point-based detectors surpass anchor-based detectors for the first time. However, they require post-processing to group key-points to the same instance, which slows down the overall computing speed. Moreover, {\emph{the boundary of an object is determined by four extreme points, the corner points and the center point are both conditional results.}} Therefore, the extreme regions have more substantial location features than the other ones.

\subsection{Dense prediction}
FPN \cite{lin2017feature} powers various detectors to achieve high-precision dense prediction, such as FCOS \cite{Tian_2019_ICCV} and FoveaBox \cite{kong2020foveabox}. In general, object detection algorithms process an image on the object level, whereas FCOS proves for the first time that the object detection task could also be solved in a per-pixel prediction fashion. This pixel-level-based detector provides a more fine-grained manner to understand an image. All these top-down methods represent the location of an object by a rectangular box. Compared with the four extreme points, such unified representation lacks the shape feature of an object, especially for those non-rigid objects with a large shape variance.

\subsection{Motivation}
Object detection involves classification and localization (bbox regression). However, there exists a misalignment between them. IoU-net \cite{jiang2018acquisition} finds that some detection results with high classification confidence have coarse bbox predictions. {\emph{Therefore, taking classification confidence as the only criterion of detection results is not accurate enough.}} BorderDet \cite{qiu2020borderdet} utilizes border features to improve detection results. It also reveals that \emph{the most important features for localization lie in the extreme point regions.}

In this paper, we provide \textbf{\emph{EPP-Net}}, a simple yet effective fully convolutional one-stage object detection method, which densely predicts the relative displacement vector between each location and the four extreme points, as shown in Fig. \ref{example1}. 
We also propose a new evaluation metric, namely, Extreme Intersection over Union ($EIoU$), to measure the similarity between two groups of extreme points, and a new loss function, namely, Extreme $IoU$ loss ($EIoU$ loss), tailored for this model. Moreover, we propose a new branch to predict the $EIoU$ between the extreme points and the matched ground-truth with the $EIoU$ servers as the localization confidence for each prediction result. By combining the predicted $EIoU$ with the classification confidence as the ranking keyword in non-maximum suppression (NMS), we show a considerable improvement in the detection results.

In summary, the contributions of this paper are as follows:
\begin{enumerate}
    \item EPP-Net decomposes the detection task into extreme points prediction and classification. Compared with the bottom-up methods, EPP-Net does not need a subsequent grouping process. 
    \item We propose $EIoU$, a normalized and scale-invariant evaluation metric, to measure the similarity between any two groups of extreme points. By incorporating $EIoU$ as the regression loss, namely, $EIoU$ loss, the accuracy with $EIoU$ loss can easily exceed that of Smooth-$\ell_{1}$ loss by 1.4\% without fine-tuning. 
    \item We present an $EIoU$ predictor to solve the misalignment problem between localization and classification. The predicted $EIoU$ serves as the localization confidence, and it is combined with the classification confidence as the ranking keyword in NMS. After appending this branch, the AP is improved by 0.5\%.   

\end{enumerate} 

\begin{figure*}[htbp]
    \centerline{\includegraphics[width=0.95\textwidth]{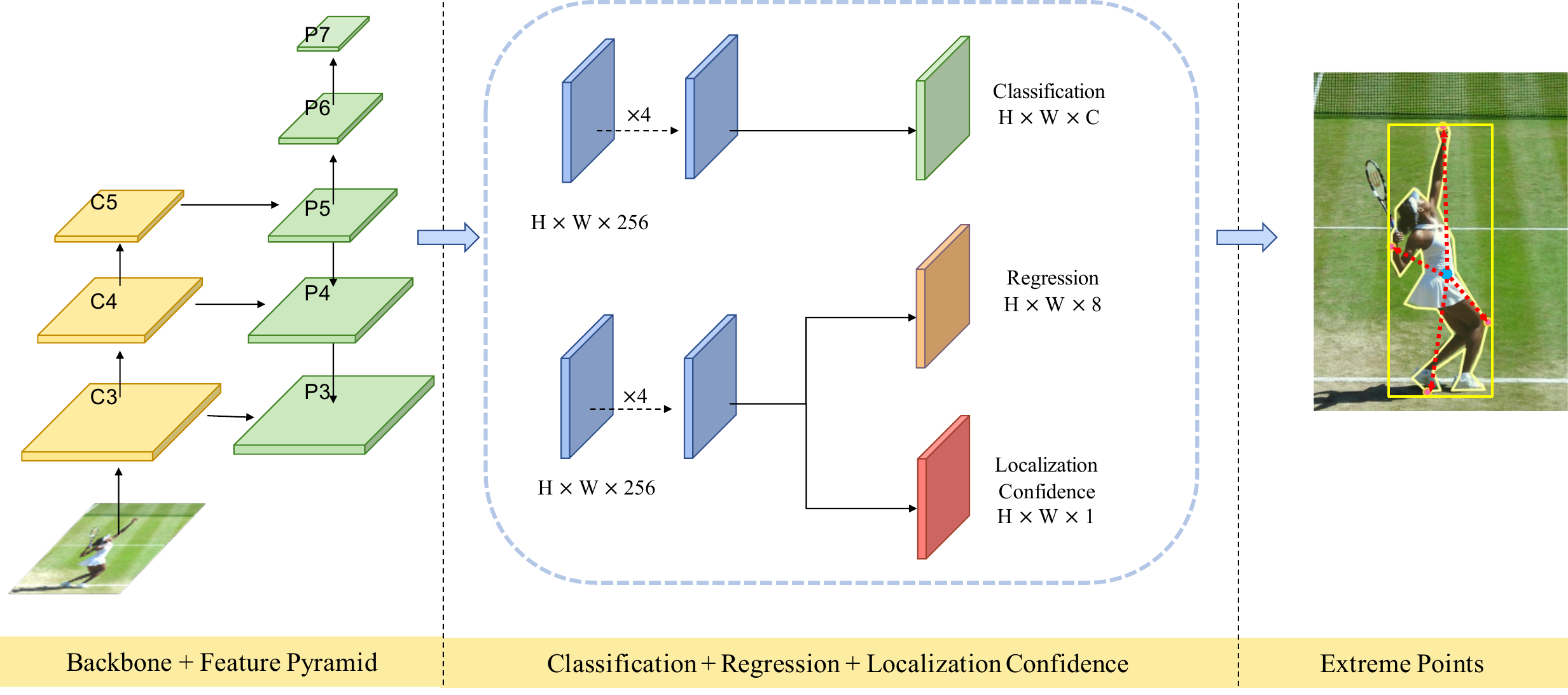}}
    \caption{\textbf{{Architecture of EPP-Net}}. This network consists of a backbone, a feature pyramid network, and two subnets, corresponding to classification and regression. For each pixel, EPP-Net outputs classification confidence, regression results, and the localization confidence.}
    \label{network} 
    \end{figure*}
\section{Related Work} 

\subsection{Anchor-Free Object Detection}
Current anchor-free detectors can achieve the same accuracy as anchor-based ones, with fewer hyperparameters and no complicated $IoU$ calculations. Despite the fact that DenseBox \cite{huang2015densebox} and YOLOv1 \cite{redmon2016you} are the earliest explorations of anchor-free models, DenseBox is not suitable for generic object detection, and the YOLO family added the anchor strategy in its subsequent versions. Therefore, these two methods are not included in the following discussions. 
\subsubsection{Key-point-based prediction}
Key-point-based detectors detect an object as one or several key-points and utilize post-processing methods to group the key points. CornerNet outputs the heatmaps of the top-left and bottom-right corners and an embedding vector for each key-point. In its grouping process, embeddings that have smaller Euclidean distances are grouped as the same instance. Based on CornerNet, \cite{duan2019centernet} adds center point prediction. In its grouping process, it also uses embedding vectors to group points. Each predicted bbox has a predefined central region and will be preserved only when the center point falls in this region. ExtremeNet predicts four extreme points and a center point for each object. In its grouping process, it uses a brute force method to enumerate all possible combinations. The box will be preserved only when the geometric center of the extreme points has a high response in the center point heatmap. The time complexities of these post-processing methods are $O(n^2)$, $O(n^2)$, and $O(n^4)$, respectively, which slow down the overall computing speed. Our EPP-Net is a top-down method so that it does not need a grouping process.
\subsubsection{Dense prediction}
FSAF \cite{zhu2019feature} employs an extra anchor-free module on the anchor-based detector for detection and feature selection. FSAF calculates the total loss for each instance and selects the pyramid level with the minimal loss to learn the instance. 
FoveaBox predicts category-sensitive semantic maps for the object's existing possibility and the bbox for each position that potentially contains an object. Our method outperforms them without the feature selection strategy and category-sensitive semantic maps. For FCOS, each location inside the object is a potential positive sample, and it directly predicts the relative distances from the four sides of the bbox to the location. It also utilizes a center-ness branch to suppress classification results far from the center region. Compared with it, our EPP-Net combines the localization and classification confidence to select the best detection results, which is more reasonable. Moreover, Instead of regressing the four bounds of the bbox, the way EPP-Net predicts is more precise.

\subsection{Localization and Classification Spatial Misalignment}
Localization is a position-sensitive task, whereas classification is not because of its translation and scale invariance properties; that is, the position or scale change of features does not affect the classification results. Therefore, a spatial misalignment exists between them. TSD \cite{song2020revisiting} proves that localization is boundary-sensitive, whereas classification is salient-area-sensitive. IoU-net \cite{jiang2018acquisition} utilizes an extra subnet to predict the $IoU$ between the detection results and ground-truth bboxes, and takes it as the ranking keyword in NMS. In contrast to IoU-net, first, IoU-Net is a two-stage, anchor-based detector while ours is a one-stage, and anchor-free detector. Second, The predicted $IoU$ in IoU-Net is class-aware, while our $EIoU$ predictor is unrelated to classes and the IoU-guided NMS is not used. Finally, our localization predictor is very light because it is only a branch of the regression subnet, while IoU-net requires a new head that is parallel with the classification and regression heads. 


\subsection{Regression Loss}
$\ell_{n}$-norm-based losses are widely used in bbox regression. However, they suffer from the scale imbalance problem, which means the loss value is affected by the scale of the bbox. $IoU$ is an evaluation metric that measures the overlap between two bboxes. \cite{yu2016unitbox} proposes $IoU$ loss based on this metric, which also inherits $IoU$'s scale invariance. When the two bboxes do not overlap, $IoU$ becomes 0 and cannot be optimized. Therefore, $GIoU$ loss \cite{rezatofighi2019generalized} is proposed to solve this problem. Standing on the shoulders of giants, we propose $EIoU$ loss to measure the similarity of two convex quadrilaterals.

\section{Method}
In this section, we briefly introduce the details of EPP-Net. We use FCOS from mmdetection \cite{chen2019mmdetection} as the baseline and ResNet-50 as the basic backbone. In EPP-Net, an object is detected as four extreme points (leftmost, top, rightmost, and bottom) by predicting the relative displacement vector in a per-pixel prediction fashion. We propose $EIoU$ as well as $EIoU$ loss for extreme point regression. Finally, we propose a novel $EIoU$ predictor for accurate key-point prediction.

\subsection{Positive Sampling with Dynamic Radius}\label{dr}
\begin{figure}[t]
        \subfigure[center sampling]{
        \begin{minipage}[t]{0.5\textwidth}
        \includegraphics[width=0.9\textwidth]{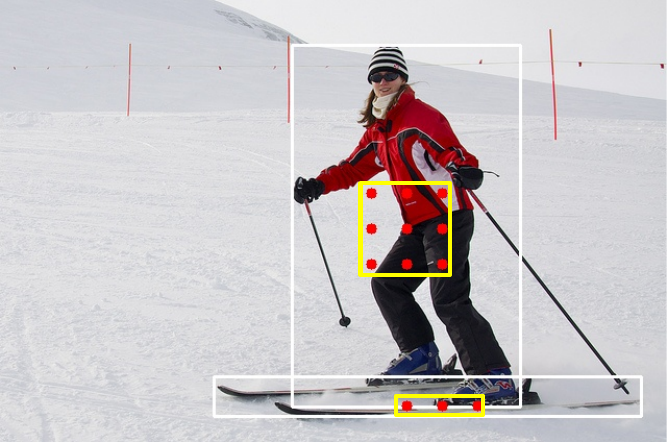}
        \end{minipage}%
        }%
        \subfigure[ours]{ 
        \begin{minipage}[t]{0.5\textwidth}
        \includegraphics[width=0.9\textwidth]{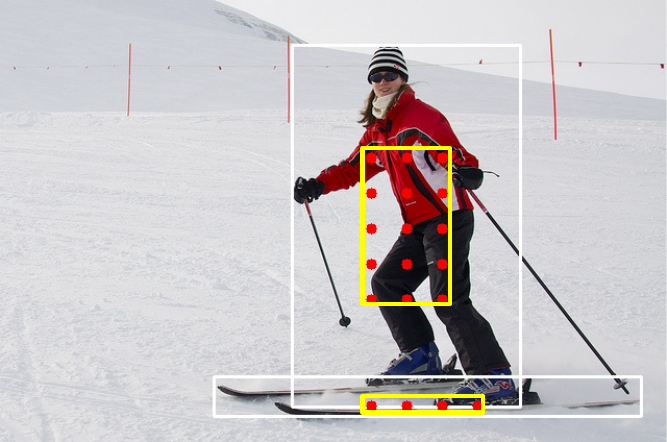}
        \end{minipage}%
        }%
        \caption{\textbf{{Positive Sampling.}} The red points denote the positive samples. The center sampling strategy from FCOS takes the positive area as a square, whereas we dynamically adjust the sampling area according to the bbox shape.}
        \label{positive}
        \end{figure}

The extreme points ground-truth is defined as $E$, where $E=\left(e_{xi}, e_{yi}\right) \in \mathbb{R}^{8}, i\in \{l,t,r,b\}$. Given a location $(x,y)$, if it falls into the target area of the ground-truth box, it is considered as a positive sample; otherwise, a negative sample. Let $(c_{x},c_{y})$ be the center point of the ground-truth box, and $s_{j}$ \cite{Tian_2019_ICCV} be the stride of feature map $j$. The target area is defined as $(c_{x}-s_{j}\ast r_{x},c_{y}-s_{j}\ast r_{y},c_{x}+s_{j}\ast r_{x},c_{y}+s_{j}\ast r_{y})$. $r_{x}$ and $r_{y}$ are the horizontal and vertical sampling radii, respectively. Considering the large difference of aspect ratio of different objects, it is improper if the sampling radii of different directions are set to be the same length. Therefore, we dynamically adjust the radius according to the aspect ratio, with the sampling radius on the longer side set to be larger, as shown in Fig. \ref{positive}(b). Let $f=\frac{w}{h}$, where $w$ and $h$ are the width and height of the ground-truth box, respectively. $r_{x}$ and $r_{y}$ are defined as follows:\\ 
\begin{equation}
    \left(r_{x}, r_{y}\right)=\left\{\begin{aligned}(1.5 \ast f, 1.5), & f>1 \\\left(1.5, \frac{1.5}{f}\right), & f<1 \end{aligned}\right.
    \end{equation}

\subsection{Network Outputs}
As shown in Fig. \ref{network}, the classification subnet outputs the classification confidence with a shape as  $H* W* C$, where $C$ is the number of MS-COCO categories \cite{lin2014microsoft}. The $C$ channels of the classification outputs correspond to $C$ binary classifiers. 

The regression subnet consists of two branches, which output the $EIoU$ prediction results and the relative displacement vector, respectively, and their shapes are $H* W* 1$ and $H* W* 8$, respectively. Details of the $EIoU$ predictor are in Chapter \ref{EIOU_Ptr}. Given a positive sample $P({p}_{x},{p}_{y})$ and the four extreme points coordinates, the relative displacement vector is $({e}_{xi}-p_{x}, {e}_{yi}-p_{y}), i\in \{l,t,r,b\}$. 

\subsection{EIoU loss}
\begin{figure}[t]
    \centerline{\includegraphics[width=0.99\textwidth]{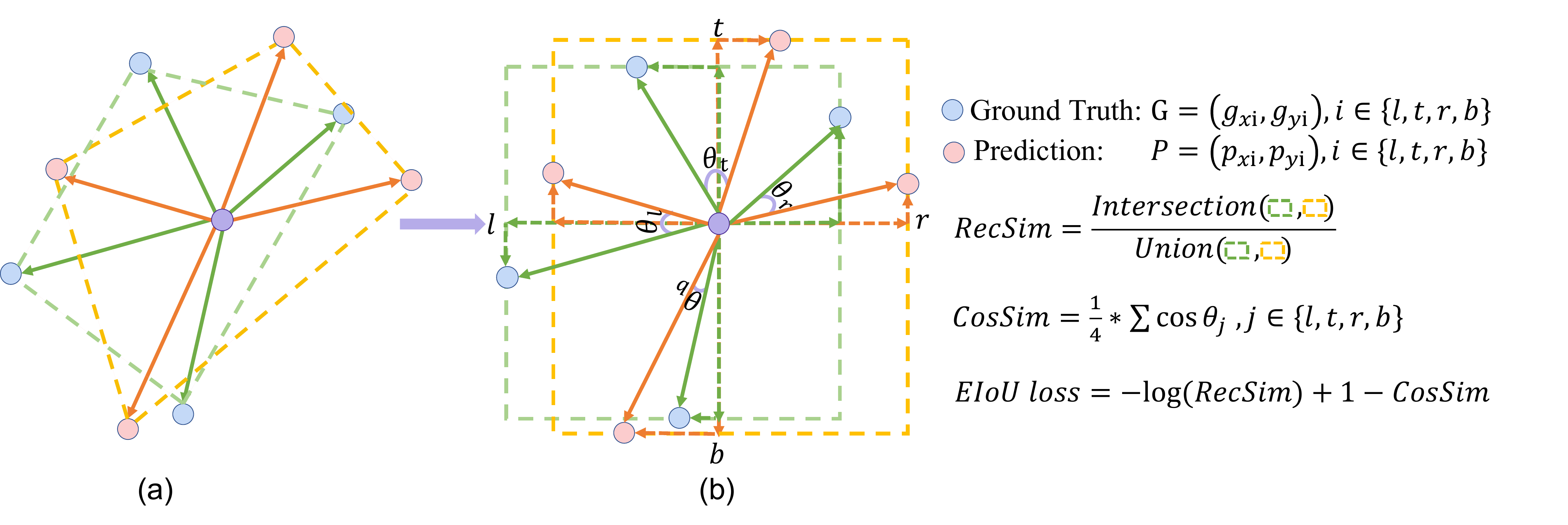}}
    \caption{\textbf{{Illustration of EIoU loss}}. The four extreme points are taken as a convex quadrilateral composed of four vectors. To simplify the calculation, the $IoU$ of the smallest enclosing rectangles and the cosine similarity between each paired vectors are used to measure the similarity of two groups of extreme points. }
    
    \label{eiou_f}
    \end{figure}

$\ell_{n}$-norm-based losses have the scale imbalance problem. Moreover, a gap exists between the $\ell_{n}$-norm and the evaluation metric $IoU$. The performances of $IoU$ loss and $GIoU$ loss prove the effectiveness of utilizing $IoU$ in regression loss. Compared with $IoU$ loss, $GIoU$ loss can optimize cases where bboxes have no overlap area. Therefore, we want to design a regression loss that inherits the scale-invariant property of $IoU$ and can compare any two convex quadrilaterals, even for non-overlapping cases. 

As shown in Fig. \ref{eiou_f} (a), the four extreme points form an irregular convex quadrilateral. Thus, calculating the $IoU$ of these two quadrilaterals seems to be optimal. However, the calculation of $IoU$ with respect to non-axis-aligned quadrilaterals is very complicated. Therefore, we choose a compromise way to simplify the calculation. The features of a quadrilateral can be decomposed into position, scale, and shape. To compare the first two features, we calculate the similarity between the two smallest enclosing rectangles of these quadrilaterals (The dotted rectangles in Fig. \ref{eiou_f} (b)), and the similarity ($RecSim$) is defined as the  $IoU$ between them. For the last feature, we use the mean value of the cosine similarity ($CosSim$) between each paired vectors to represent the overall shape difference, as shown in Equation \ref{CosSim}. The cosine similarity is equivalent to the angle between vectors, thus perfectly reflecting the shape difference. 
\begin{equation}
    CosSim= \frac{1}{4} * \sum \cos \theta_{j}, j \in\{l, t, r, b\}
\label{CosSim}
\end{equation}

Therefore, the similarity of any two convex quadrilaterals on the Euclidean plane can be measured by $EIoU$. If not specified, we use the $IoU$ between the two smallest enclosing rectangles as the $RecSim$ in all equations. The definition of $EIoU$ is shown in Equation \ref{EIOU_q}. 
\begin{equation}
    EIoU= \frac{1}{2} *(IoU+\frac{1+CosSim}{2})
    \label{EIOU_q}
    \end{equation}

The properties of $EIoU$ are as follows:
\begin{enumerate}
    \item $IoU$ and cosine similarity are scale-invariant. Thus, $EIoU$ also inherits this property. 
    \item For any two convex quadrilaterals A and B. $A, B \subseteq \mathbb{S}, 0 \leq {IoU}(A, B) \leq 1, -1 \leq {CosSim}(A, B) \leq 1 $, that $0 \leq {EIoU}(A, B) \leq 1$ can be easily obtained. Therefore, $EIoU$ is an normalized evaluation metric.
    \item $IoU$ can be considered a special case of $EIoU$. When both convex quadrilaterals are axis-aligned rectangles, $EIoU$ is equivalent to $IoU$.
    \end{enumerate}


    \begin{algorithm}[t]
        \caption{$EIoU$ loss Forward} 
        \label{alg:EIoU_forward}
        \textbf{Input}: $\mathrm{G}=\left(g_{x \mathrm{i}}, g_{y \mathrm{i}}\right), i \in\{l, t, r, b\} $ as the ground-truth.\\
        \textbf{Input}: $\mathrm{P}=\left(p_{x \mathrm{i}}, p_{y \mathrm{i}}\right), i \in\{l, t, r, b\} $ as the prediction.\\
        \textbf{Input}: $\{{\theta}_{l},{\theta}_{t},{\theta}_{r},{\theta}_{b}\}$ the angles between paired vectors.\\
        \textbf{Output}: $\mathcal{L}_{EIoU}$ 
        \begin{algorithmic}[1] 
        \FOR{ each prediction}
        \STATE ${CosSim}=\frac{1}{4} * \sum \cos \theta_{j}, j \in\{l, t, r, b\}$
        \STATE $\begin{aligned}
            &\mathcal{I}_{x1}=\max \left({g}_{xl},{p}_{xl}\right),
            \mathcal{I}_{y1}=\max \left({g}_{yt},{p}_{yt}\right) 
            \end{aligned}$  
        \STATE$\begin{aligned}
                &\mathcal{I}_{x2}=\min \left({g}_{xr},{p}_{xr}\right),
                \mathcal{I}_{y2}=\min \left({g}_{yb},{p}_{yb}\right)
                \end{aligned}$ 
        \STATE $\begin{aligned}
            &{\mathcal{A}_{g}}=({g}_{xr}-{g}_{xl})*
            ({g}_{yb}-{g}_{yt})
            \end{aligned}$
        \STATE$\begin{aligned}
            &{\mathcal{A}_{p}}=({p}_{xr}-{p}_{xl})*
            ({p}_{yb}-{p}_{yt})
            \end{aligned}$
        \STATE $\begin{aligned}
            \mathcal{I}=(\mathcal{I}_{x2}-\mathcal{I}_{x1})*
            (\mathcal{I}_{y2}-\mathcal{I}_{y1})
        \end{aligned}$
        \IF {$\begin{aligned}\mathcal{I}>0\end{aligned}$} 
        \STATE $\begin{aligned}
            \mathcal{U}={\mathcal{A}_g}+\mathcal{A}_p-\mathcal{I}
        \end{aligned}$
        \STATE $\begin{aligned}
            IoU= \max \left(\frac{\mathcal{I}}{\mathcal{U}},e^{-6}\right)
        \end{aligned}$
        \STATE $\begin{aligned}
            \mathcal{L}_{EIoU}=-ln(IoU)+(1- {CosSim})
        \end{aligned}$
        \ELSE 
        \STATE $\begin{aligned}
            \mathcal{L}_{EIoU}=-ln(e^{-6})+(1- {CosSim})
        \end{aligned}$
        \ENDIF 
        \ENDFOR
        \end{algorithmic}
        \end{algorithm}

With $IoU$ ranges between 0 and 1, the cross-entropy of $IoU$ is $-1*ln(IoU)$. The range of cosine similarity is between -1 and 1 and the cosine similarity difference is defined as $1-CosSim$. Therefore, $\mathcal{L}_{EIoU}$ is defined as Equation \ref{EIOU_loss}:

\begin{equation}
    \begin{aligned}
        \mathcal{L}_{EIoU} &=\mathcal{L}_{RecSim}+\mathcal{L}_{CosSim} \\
    &=-\ln (I o U)+(1-CosSim)
    \end{aligned}
    \label{EIOU_loss}
    \end{equation}

The details of $EIoU$ loss is shown in Algorithm \ref{alg:EIoU_forward}. $EIoU$ loss has the following properties:

\begin{enumerate}
    \item $EIoU$ loss is invariant to scale changes.
    \item The value range of $EIoU$ loss is $[0,8]$. Its value will become 0 only when the two groups of extreme points completely coincide; otherwise, it will be positive. Consequently, $EIoU$ loss can optimize any two groups of extreme points. 
    \end{enumerate}

\subsection{EIoU Predictor}\label{EIOU_Ptr}
Here, we provide this $EIoU$ predictor to deal with the misalignment problem between localization and classification. Object detection methods usually predict many bboxes with large overlapping areas. Therefore, the NMS algorithm is used to filter out poor prediction results with the classification confidence as the ranking keyword. However, this method may filter out the detection results with good bbox predictions but low classification confidence. Thus, our $EIoU$ predictor scores each regression result by predicting the $EIoU$ between each predicted bbox and its associated ground-truth. By doing so, we take the localization and classification confidence together as the evaluation criteria for prediction results.

During inference, we multiply the classification confidence and the $EIoU$ prediction results as the final ranking keyword in NMS, as shown in Equation \ref{ranking}.

\begin{equation}
    \text {ranking}={EIoU} * \text {cls-confidence}
    \label{ranking}
    \end{equation}
 
\subsection{Optimization}
The total loss of this model is formulated as follows:
\begin{equation}
    \begin{aligned}
        \mathcal{L}=\lambda_{cls} \mathcal{L}_{cls}+\lambda_{reg} \mathcal{L}_{reg}
        +\lambda_{\mathrm{eioup}} \mathcal{L}_{eioup}
    \end{aligned}
    \end{equation}

$\mathcal{L}_{cls}$ is focal loss for classification as in \cite{lin2017focal}, and $\mathcal{L}_{reg}$ is defined in Equation \ref{EIOU_loss}. $\mathcal{L}_{eioup}$ is BCE loss for $EIoU$ predictions. To balance losses of all subtasks, hyperparameters $\lambda_{cls}$, $\lambda_{reg}$, and $\lambda_{eioup}$ are all set as 1.

\section{Experiments}
In this section, we perform several experiments on the MS-COCO dataset \cite{lin2014microsoft} to show the effectiveness of EPP-Net and its counterparts. EPP-Net is trained on the COCO train2017 split (115K images) and evaluated on the COCO val2017 split (5K images) for the ablation study. Visualization experiments are also conducted on the val2017 split. We also upload the detection results on the test-dev split (20K images) with different backbones to the MS-COCO server to compare our EPP-Net with recent state-of-the-art detectors.
\subsection{Implementation Details}
Our implementation is based on mmdetection \cite{chen2019mmdetection} with Pytorch 1.6. Extreme points are computed from the polygonal mask annotations following the extraction strategy from \cite{zhou2019bottom}. The hyperparameters in our model follow those in FCOS, and we use pre-trained models on ImageNet to initialize network weights. If not specified, we use ResNet-50 and feature pyramid network as our basic network. We train this network with stochastic gradient descent and a total batch size of 16 images on 8 NVIDIA TITAN RTX GPUs for 90K iterations. We set the initial learning rate as 0.01, and the momentum and the weight decay as 0.9 and 0.0001, respectively. We decrease the learning rate by 10 at epochs 8 and epoch 11. The $IoU$ threshold in NMS is set as 0.6.
\subsection{Ablation Study}
We perform several groups of ablation experiments to validate the effectiveness of different counterparts. All test results are reported on MS-COCO val2017 split.
\begin{figure*}[t]
    \centering
    \subfigure[FCOS]{
        \begin{minipage}[t]{0.99\textwidth}
        \includegraphics[width=0.99\textwidth]{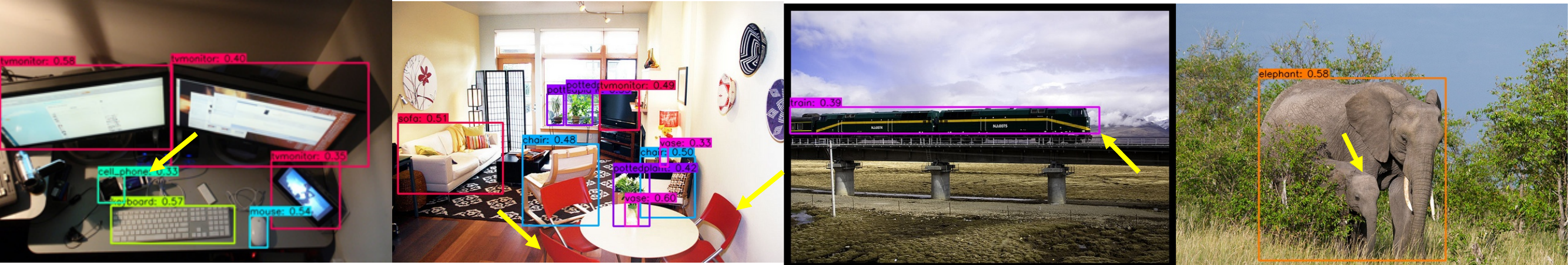}
        \end{minipage}%
        }%
        
        \subfigure[EPP-Net]{ 
        \begin{minipage}[t]{0.99\textwidth}
        \includegraphics[width=0.99\textwidth]{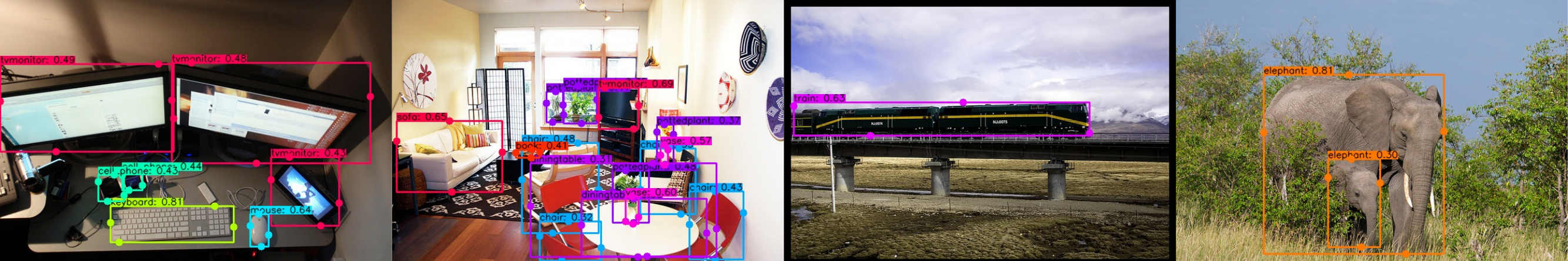}
        \end{minipage}%
        }%
        \caption{\textbf{{Qualitative results on the val2017 split}}. Extreme points and bbox detection results of EPP-Net are shown on the same image. With ResNet-50, our model (The model with AP 39.5\%) can achieve excellent detection results in various scenes.}
    \label{imgdemo} 
    \end{figure*} 
    \begin{table}[b]
        \caption{\textbf{{EPP-Net \emph{vs.} FCOS.}} 
        Comparisons on the val2017 split with ResNet-50-FPN as the backbone. 
        ``bbox'' and ``ex'': representing objects by bounding boxes and extreme points.``loc'': the localization confidence branch. ``ctr-ness'' and ``ctr'': center-ness and center sampling in FCOS. ``dr'': our positive sampling strategy, details are in section \ref{dr}.}
        \centering
        {\begin{tabular}{l|ccc|c|cc|ccc}
        \toprule
        Method  &${Reg}$&loc&sampling&AP&$\mathrm{AP}_{50}$&$\mathrm{AP}_{75}$&$\mathrm{AP}_{S}$&$\mathrm{AP}_{M}$&$\mathrm{AP}_{L}$\\ 
        \hline
        FCOS\cite{Tian_2019_ICCV}    &$\mathcal{L}_{GIoU}+$bbox&ctr-ness&ctr&38.6&57.4&41.4&22.3&42.5&49.8     \\
        \hline         
        EPP-Net       &$\mathcal{L}_{EIoU}+$ex&ctr-ness&ctr&{38.9}&{57.3}&{42.2}&{23.0}&{42.7}&{50.0}\\
        EPP-Net       &$\mathcal{L}_{EIoU}+$ex&$EIoU$&ctr&{39.4}&{57.7}&\textbf{43.2}&\textbf{23.4}&\textbf{43.6}&{50.6}\\
        EPP-Net       &$\mathcal{L}_{EIoU}+$ex&$EIoU$&dr&\textbf{39.5}&\textbf{58.1}&{42.9}&{23.1}&{43.4}&\textbf{51.1}\\
                    
        \bottomrule
        \end{tabular}}
        
        \label{R50_map}
        \end{table}
\subsubsection{Overall performance}
We compare our method with FCOS to evaluate the overall performance of EPP-Net. We use the control variable method to validate each counterpart in EPP-Net, which are $EIoU$ loss, the $EIoU$ predictor, and our positive sampling strategy. As shown in Table \ref{R50_map}, the best model of EPP-Net outperforms FCOS with an AP of 39.5\%. Compared with bounding boxes, the representation of extreme points can improve the AP by 0.3\% (The model with AP 38.9).   Our localization branch improves the AP@75 by 1.0\%, which indicates the effectiveness of the $EIoU$ predictor. With our positive sampling strategy, we observe a considerable improvement of AP in large objects, namely, 0.5\%. We explain as follows: First, large objects are more likely to sample more true positive samples. Second, the shape variance of large objects is larger than that of small objects, therefore, a dynamic sampling radius following the changes of object shapes is more appropriate. The visualization of the detection results is shown in Fig. \ref{imgdemo} and one can see that our detection results are more accurate than that of FCOS.

\begin{table}[t]
    \caption{\textbf{{EIoU loss \emph{vs.} Smooth-$\ell_{1}$ loss.}} Settings are the same as the EPP-Net in Table \ref{R50_map}. The performance of $EIoU$ loss is much better than that of Smooth-$\ell_{1}$ loss.}
    \centering
    {\begin{tabular}{l|c|cc|ccc} 
    \toprule
    $loss$&AP&$\mathrm{AP}_{50}$&$\mathrm{AP}_{75}$&$\mathrm{AP}_{S}$&$\mathrm{AP}_{M}$&$\mathrm{AP}_{L}$\\ 
    \hline
    ${EIoU}$&\textbf{39.5}&\textbf{58.1}&\textbf{42.9}&\textbf{23.1}&\textbf{43.4}&\textbf{51.1}\\ 
    w/ Smooth-$\ell_{1}$&38.1&57.5&40.7&21.5&42.2&49.9\\  
    \bottomrule
    \end{tabular}}
    
    \label{EIOU_loss_compare}
    \end{table}
\subsubsection{EIoU loss} 
The IoU-based losses require a 4D vector to represent the object which is incompatible with our regression task (8D vector). Therefore, we take Smooth-$\ell_{1}$ loss and $EIoU$ loss as the regression loss, respectively, to prove the effectiveness of $EIoU$ loss. The results are shown in Table \ref{EIOU_loss_compare}. Smooth-$\ell_{1}$ loss achieves an AP of 38.1\%, and our $EIoU$ loss outperforms it by 1.4\%. $\mathrm{AP}_{S}$, $\mathrm{AP}_{M}$, and $\mathrm{AP}_{L}$ are all raised considerably, which proves the importance of the scale invariance property of regression loss.
\begin{table}[b]
    \caption{\textbf{{EIoU \emph{vs.} other counterparts.}} EIoU-branch denotes our $EIoU$ predictor. QFL denotes the joint representation of $IoU$ score and classification.}
    \centering
    {\begin{tabular}{l|c|cc|ccc} 
    \toprule
    
    Type&AP&$\mathrm{AP}_{50}$&$\mathrm{AP}_{75}$&$\mathrm{AP}_{S}$&$\mathrm{AP}_{M}$&$\mathrm{AP}_{L}$  \\
    \hline
    EIoU-branch&\textbf{39.5}&\textbf{58.1}&\textbf{42.9}&\textbf{23.1}&\textbf{43.4}&\textbf{51.1}\\
    centerness-branch \cite{Tian_2019_ICCV}&38.6&57.4&41.4&22.3&42.5&49.8\\
    IoU-branch \cite{jiang2018acquisition,wu2020iou}&{38.7}&{56.7}&{42.0}&{21.6}&{43.0}&{50.3}\\
    QFL \cite{li2020generalized}&{39.0}&{57.8}&{41.9}&{22.0}&{43.1}&{51.0}\\ 
    \bottomrule
    \end{tabular}}  
    \label{EIOU_Ctr}
    \end{table}
\subsubsection{EIoU predictor} 
As shown in Table \ref{EIOU_Ctr}, we compare our localization confidence predictor with other strategies. The center-ness in FCOS is a predefined localization confidence heatmap with the belief that the center area predicts better localization results. However, the geometric center of some objects does not fall in the foreground area, such as the crescent moon. Compared with center-ness, taking $IoU$ or $EIoU$ as the localization confidence is more generalized and has achieved better performance. The IoU-branch in IoU-Net and QFL in GFocal loss are class-aware, while our $EIoU$ branch is independent of classes. Our $EIoU$ outperforms all other counterparts with an AP of 39.5\%. We can conclude that the $EIoU$ predictor can improve the detection accuracy by suppressing inaccurate localization results.
\subsection{State-of-the-art Comparisons} 
Table \ref{sota} shows the comparison results between EPP-Net and state-of-the-art detectors. We use multi-scale training with the shorter side of input images randomly resized from 640 to 800 and the longer side less than 1333. The training process follows the 2 $\times$ stratedgy  in \cite{chen2019mmdetection}. Test results are reported on the MS-COCO test-dev split by uploading the detection results to the MS-COCO server. Our model achieves a substantial improvement with different backbones. Compared with anchor-based RetinaNet, our model achieves an improvement of 5.0\% in AP with backbone ResNeXt-101. EPP-Net also outperforms key-point-based detectors, CornerNet and ExtremeNet, with better accuracy and without the grouping process. Moreover, EPP-Net outperforms FCOS by 1.0\% and achieves an AP of 45.8\% with ResNeXt-101. Finally, the performance of the best model reaches 50.3\% AP with ResNeXt-101-DCN as the backbone.

\begin{table*}[t]
    \caption{\textbf{{EPP-Net \emph{vs.} State-of-the-art Detectors.}} ``\dag'' indicates the multi-scale testing and settings are the same as in \cite{tian2020fcos}. }
    \centering
    \scalebox{1.0}{ 
    {\begin{tabular}{l|c|ccc|ccc} 
    \toprule
    Method&Backbone&AP&$\mathrm{AP}_{50}$&$\mathrm{AP}_{75}$&$\mathrm{AP}_{S}$&$\mathrm{AP}_{M}$&$\mathrm{AP}_{L}$\\ 
    \hline
    \textbf{Anchor-Based}\\
    Faster R-CNN w/ FPN \cite{lin2017feature}&ResNet-101&36.2&59.1&39.0&18.2&39.0&48.2\\ 
    YOLOv4 \cite{bochkovskiy2020yolov4}&CSPDarknet-53&43.5&65.7&47.3&26.7&46.7&53.3\\
    RetinaNet \cite{lin2017focal}&ResNeXt-101&40.8&61.1&44.1&24.1&44.2&51.2\\
    IoU-Net \cite{jiang2018acquisition}&ResNet-101&40.6&59.0&-&-&-&-\\
    FSAF \cite{zhu2019feature}&ResNeXt-101&42.9&63.8&46.3&26.6&46.2&52.7\\
    ATSS\cite{zhang2020bridging}&ResNeXt-101-DCN&47.7&66.5&51.9&29.7&50.8&59.4\\
    GFL\cite{li2020generalized}&ResNeXt-101-DCN&48.2&67.4&52.6&29.2&51.7&60.2\\
    \hline
    \textbf{Anchor-Free}\\
    CornerNet \cite{law2018cornernet}&Hourglass-104&40.5&59.1&42.3&21.8&42.7&50.2\\
    ExtremeNet \cite{zhou2019bottom}&Hourglass-104&40.2&55.5&43.2&20.4&43.2&53.1\\
    CenterNet-HG \cite{zhou2019objects}&Hourglass-104&42.1&61.1&45.9&24.1&45.5&52.8\\
    CenterNet511 \cite{duan2019centernet}&Hourglass-104&44.9&62.4&48.1&25.6&47.4&57.4\\
    RepPoints \cite{yang2019reppoints}&ResNet-101&41.0&62.9&44.3&23.6&44.1&51.7\\
    FoveaBox-align \cite{kong2020foveabox}&ResNeXt-101&43.9&63.5&47.7&26.8&46.9&55.6\\
    FCOS-imprv \cite{tian2020fcos}&ResNeXt-101&44.8&64.4&48.5&27.7&47.4&55.0\\
    FCOS-imprv\dag \cite{tian2020fcos}&ResNeXt-101-DCN&49.1&68.0&53.9&31.7&51.6&61.0\\
    \hline
    EPP-Net\dag&ResNet-50&44.0&62.2&48.6&28.3&46.5&54.0\\
    EPP-Net&ResNeXt-101&45.8&65.1&49.9&28.1&49.0&56.4\\
    EPP-Net\dag&ResNeXt-101&48.1&66.7&53.0&31.8&50.9&58.8\\
    EPP-Net&ResNeXt-101-DCN&{48.3}&{67.5}&{52.5}&{29.0}&{51.6}&{61.6}\\ 
    EPP-Net\dag&ResNeXt-101-DCN&\textbf{50.3}&\textbf{68.3}&\textbf{55.0}&\textbf{33.0}&\textbf{53.0}&\textbf{62.4}\\     
    \bottomrule
    \end{tabular}}}
    
    \label{sota}
    \end{table*}
    \section{Conclusion}
    In this paper, we present EPP-Net as a new method to detect an object by predicting the relative displacement vector between each location and the four extreme points. We also propose $EIoU$, a novel evaluation metric, to measure the similarity between two groups of extreme points. Moreover, our proposed $EIoU$ loss can deal with the scale imbalance problem, which outperforms Smooth-$\ell_{1}$ loss. Furthermore, we propose the $EIoU$ predictor, which helps the detector obtain better localization results. The detection results on the MS-COCO reveal that our method can achieve state-of-the-art accuracy. 

%
%

%
%
%
\bibliographystyle{splncs04}
\bibliography{mybibliography}
%




\end{document}